\documentclass{article}
\usepackage{proceed2e}
\usepackage[margin=1in]{geometry}

\usepackage{times}

\usepackage{graphicx}
\usepackage{booktabs} 
\usepackage{hyperref}
\usepackage{url}
\usepackage{color}
\usepackage{amstext} 
\usepackage{subcaption}
\usepackage{rotating, graphicx} 
\usepackage{multirow}
\usepackage{mathrsfs}
\usepackage{amsmath} 
\usepackage{nth}
\usepackage{amsfonts}
\usepackage{mathtools}
\usepackage{stmaryrd}
\usepackage{footnote}
\makesavenoteenv{tabular}
\makesavenoteenv{table}

\usepackage{siunitx} 
\usepackage{booktabs}
\usepackage{tabularx}
\usepackage{array}   
\usepackage{natbib}
\setcitestyle{open={(},authoryear,close={)}}
\usepackage{enumitem}
\setlist{leftmargin=4.5mm}

\DeclarePairedDelimiter\abs{\lvert}{\rvert}%

\title{GaAN: Gated Attention Networks for \\Learning on Large and Spatiotemporal Graphs}

\author{Jiani Zhang$^{1}$\thanks{\quad These two authors contributed equally.}, Xingjian Shi$^{2}$\footnotemark[1], Junyuan Xie$^{3}$, Hao Ma$^{4}$, Irwin King$^{1}$, Dit-Yan Yeung$^{2}$\\
	{\small $^1$The Chinese University of Hong Kong, Hong Kong, China,}
	\emph{\small \{jnzhang, king\}@cse.cuhk.edu.hk}\\
	{\small $^2$Hong Kong University of Science and Technology, Hong Kong, China,}
	\emph{\small \{xshiab, dyyeung\}@cse.ust.hk}\\
	{\small $^3$Amazon Web Services, WA, USA,}
	\emph{\small junyuanx@amazon.com}\\
	{\small $^4$Microsoft Research, WA, USA,}
	\emph{\small haoma@microsoft.com }
}
\begin{document}
	
	\maketitle
	
	\begin{abstract}
		We propose a new network architecture, \emph{\underline{Ga}ted \underline{A}ttention \underline{N}etworks} (GaAN), for learning on graphs. Unlike the traditional multi-head attention mechanism, which equally consumes all attention heads, GaAN uses a convolutional sub-network to control each attention head's importance. We demonstrate the effectiveness of GaAN on the inductive node classification problem. Moreover, with GaAN as a building block, we construct the \emph{Graph Gated Recurrent Unit} (GGRU) to address the traffic speed forecasting problem. Extensive experiments on three real-world datasets show that our GaAN framework achieves state-of-the-art results on both tasks.
	\end{abstract}
	
	\section{INTRODUCTION}
	
	Many crucial machine learning tasks involve graph structured datasets, such as classifying posts in a social network~\citep{hamilton2017inductive}, predicting interfaces between proteins~\citep{fout2017protein} and forecasting the future traffic speed in a road network~\citep{li2017graph}. The main difficulty in solving these tasks is how to find the right way to express and exploit the graph's underlying structural information. Traditionally, this is achieved by calculating various graph statistics like degree and centrality, using graph kernels, or extracting human engineered features~\citep{hamilton2017representation}.
	
	Recent research, however, has pivoted to solving these problems by \emph{graph convolution}~\citep{duvenaud2015convolutional,atwood2016diffusion,kipf2017semi,fout2017protein,hamilton2017inductive,velivckovic2017graph, li2017graph}, which generalizes the standard definition of convolution over a regular grid topology~\citep{gehring17a, krizhevsky2012imagenet}
	to `convolution' over graph structures.  
	The basic idea behind `graph convolution' is to develop a localized parameter-sharing operator on a set of neighboring nodes to aggregate a local set of lower-level features. We refer to such an operator as a {\em graph aggregator}~\citep{hamilton2017inductive} and the set of local nodes as the {\em receptive field} of the aggregator. Then, by stacking multiple graph aggregators, we build a deep neural network~\citep{lecun2015deep} model which can be trained end-to-end to extract the local and global features across the graph.
	Note that we use the spatial definition instead of the spectral definition~\citep{hammond2011wavelets, bruna2014spectral} of graph convolution because the full spectral treatment requires eigendecomposition of the Laplacian matrix, which is computationally intractable on large graphs, while the localized versions~\citep{defferrard2016convolutional,kipf2017semi} can be interpreted as graph aggregators~\citep{hamilton2017inductive}.
	
	Graph aggregators are the basic building blocks of graph convolutional neural networks. A model's ability to capture the structural information of graphs is largely determined by the design of its aggregators. Most existing graph aggregators are based on either pooling over neighborhoods~\citep{kipf2017semi, hamilton2017inductive} or computing a weighted sum of the neighboring features~\citep{monti2017geometric}. In essence, functions that are permutation invariant and can be dynamically resizing are eligible graph aggregators. One class of such functions is the neural attention network~\citep{bahdanau2014neural}, which uses a subnetwork to compute the correlation weight of the elements in a set. Among the family of attention models, the multi-head attention model has been shown to be effective for machine translation tasks~\citep{lin2017structured,vaswani2017attention}. It has later been adopted as a graph aggregator to solve the node classification problem~\citep{velivckovic2017graph}. 
	A single attention head sums the elements that are similar to the query vector in one representation subspace. Using multiple attention heads allows exploring features in different representation subspaces, which can provide more modeling power in nature. However, treating each attention head equally loses the opportunity to benefit from some attention heads which are inherently more important than others.
	
	To this end, we propose the \emph{\underline{Ga}ted \underline{A}ttention \underline{N}etworks} (GaAN) for learning on graphs. GaAN uses a small convolutional subnetwork to compute a soft gate at each attention head to control its importance. 
	Unlike the traditional multi-head attention that admits all attended contents, the gated attention can modulate the amount of attended content via the introduced gates. Moreover, since only a simple and light-weighted subnetwork is introduced in constructing the gates, the computational overhead is negligible and the model is easy to train. We demonstrate the effectiveness of our new aggregator by applying it to the inductive node classification problem. We also improve the sampling strategy introduced in~\citep{hamilton2017inductive} to reduce the memory cost and increase the run-time efficiency, in order to train our model and other graph aggregators on relatively large graphs. Furthermore, since our proposed aggregator is very general, we extend it to construct a \emph{Graph Gated Recurrent Unit} (GGRU), which is directly applicable for spatiotemporal forecasting problem. Extensive experiments on two node classification datasets, PPI and Reddit~\citep{hamilton2017inductive}, and one traffic speed forecasting dataset, METR-LA~\citep{li2017graph}, show that GaAN consistently outperforms the baseline models and achieves the state-of-the-art performance.

	In summary, our main contributions include: (a) a new multi-head attention-based aggregator with additional gates on the attention heads; (b) a unified framework for transforming graph aggregators to graph recurrent neural networks; and (c) the state-of-the-art prediction performance on three real-world datasets.

	\section{NOTATIONS}
	We denote vectors with bold lowercase letters, matrices with bold uppercase letters and sets with calligraphy letters. We denote a single fully-connected layer with a non-linear activation $\alpha(\cdot)$ as $\text{FC}_\theta^{\alpha}(\mathbf{x}) = \alpha(\mathbf{W}\mathbf{x} +\mathbf{b})$, where $\theta = \{\mathbf{W}, \mathbf{b}\}$ are the parameters.
	Also, $\theta$ with different subscripts mean different transformation parameters.
	For activation functions, we denote $h(\cdot)$ to be the LeakyReLU activation~\citep{xu2015empirical} with negative slope equals to 0.1 and $\sigma(\cdot)$ to be the sigmoid activation. $\text{FC}_\theta(\mathbf{x})$ means applying no activation function after the linear transform. We denote $\oplus$ as the concatenation operation and $\bigparallel_{k=1}^K \mathbf{x}_k$ as sequentially concatenating $\mathbf{x}_1$ through $\mathbf{x}_K$. We denote the Hadamard product as `$\circ$' and the dot product between two vectors as $\langle \cdot, \cdot \rangle$. 
	
	\section{RELATED WORK}
	In this section, we will review relevant research on learning on graphs. Our model is also related to many graph aggregators proposed by previous work. We will discuss these aggregators in Section~\ref{sec:existing_aggregator}.

	\textbf{Neural attention mechanism}\quad
	Neural attention mechanism is widely adopted in deep learning literature and many variants have been proposed~\citep{chorowski2014end, xu2015show, seo2016bidirectional, vaswani2017attention}. Among them, our model takes inspiration from the multi-head attention architecture proposed in~\citep{vaswani2017attention}. Given a query vector $\mathbf{q}$ and a set of key-value pairs $\{(\mathbf{k}_1, \mathbf{v}_1), ..., (\mathbf{k}_n, \mathbf{v}_n)\}$, a single attention head computes a weighted combination of the value vectors $\sum_{i=1}^n w_i \textbf{v}_i$. The weights are generated by applying softmax to the inner product between the query and keys, i.e., $\mathbf{w} = \text{softmax}(\{\mathbf{q}^T\mathbf{k}_1, ...,  \mathbf{q}^T\mathbf{k}_n\})$. In the multi-head case, the outputs of $K$ different heads are concatenated to form an output vector with fixed dimensionality. The difference between the proposed model, GaAN, and the multi-head attention mechanism is that we compute additional gates to control the importance of each head's output.

	\textbf{Graph convolutional networks on large graph}\quad
	Applying graph convolution on large graphs is challenging because the memory complexity is proportional to the total number of nodes, which could be hundreds of thousands of nodes in large graphs~\citep{hamilton2017inductive}. To reduce memory usage and computational cost, \citep{hamilton2017inductive} proposed the GraphSAGE framework that uses a sampling algorithm to select a small subset of the nodes and edges. On each iteration, GraphSAGE first uniformly samples a mini-batch of nodes. Then, for each node, only a fixed number of neighborhoods are selected for aggregation. 
	More recently, Chen et al.~\citep{chen2018fastgcn} proposed a new sampling method that randomly samples two sets of nodes according to a proposed distribution. However, this method is only applicable to one aggregator, i.e., the \emph{Graph Convolutional Network} (GCN)~\citep{kipf2017semi}.

	\textbf{Graph convolution networks for spatiotemporal forecasting}\quad
	Recently, researchers have applied graph convolution, which is commonly used for learning on static graphs, to spatiotemporal forecasting. \citep{seo2016structured} proposed \emph{Graph Convolutional Recurrent Neural Network} (GCRNN), which replaced the fully-connected layers in LSTM~\citep{hochreiter1997long} with the ChebNet operator~\citep{defferrard2016convolutional}, and applied it to a synthetic video prediction task. 
	Li et al.~\citep{li2017graph} proposed \emph{Diffusion Convolutional Recurrent Neural Network} (DCRNN) to address the traffic forecasting problem, where the goal is to predict future traffic speeds in a sensor network given historic traffic speeds and the underlying road graph. DCRNN replaces the fully-connected layers in GRU~\citep{chung2014empirical} with the diffusion convolution operator~\citep{atwood2016diffusion}. Furthermore, DCRNN takes the direction of graph edges into account. The difference between our GGRU with GCRNN and DCRNN is that we have proposed a unified method for constructing a recurrent neural network based on an arbitrary graph aggregator rather than proposing a single model.

	\section{GATED ATTENTION NETWORKS}
	In this section, we first give a generic formulation of graph aggregators followed by the multi-head attention mechanism. Then, we introduce the proposed gated attention aggregator. Finally, we review the other kinds of graph aggregators proposed by previous work and explain their relationships with ours.
	
	\textbf{Generic formulation of graph aggregators} \quad
	Given a node $i$ and its neighboring nodes $\mathcal{N}_i$, a graph aggregator is a function $\gamma$ in the form of $\mathbf{y}_i = \gamma_{\Theta}(\mathbf{x}_i, \{\mathbf{z}_{\mathcal{N}_i}\})$,
	where $\mathbf{x}_i$ and $\textbf{y}_i$ are the input and output vectors of the center node $i$. $\mathbf{z}_{\mathcal{N}_i} = \{\mathbf{z}_j | j \in \mathcal{N}_i\}$ is the set of the reference vectors in the neighboring nodes and $\Theta$ is the learnable parameters of the aggregator. In this paper, we do not consider aggregators that use edge features. However, it is straightforward to incorporate edges in our definition by defining $\mathbf{z}_j$ to contain the edge feature vectors $\mathbf{e}_{i, j}$.

	\subsection{MULTI-HEAD ATTENTION AGGREGATOR}
	We linearly project the center node feature $\mathbf{x}_i$ to get the query vector and project the neighboring node features to get the key and value vectors. We then apply the multi-head attention mechanism~\citep{vaswani2017attention} to get the final aggregation function. The detailed formulation of the multi-head attention aggregator is as follows:
	\vskip -0.1in
	\begin{equation}
	\begin{aligned}
	& \mathbf{y}_i = \text{FC}_{\theta_o}(\mathbf{x}_i \oplus \bigparallel_{k=1}^K  \sum_{j\in \mathcal{N}_i}w^{(k)}_{i, j}\text{FC}^h_{\theta_{v}^{(k)}}(\mathbf{z}_j)),\\
	& w^{(k)}_{i, j} = \frac{\exp(\phi_w^{(k)}(\mathbf{x}_i,\mathbf{z}_j))}{\sum_{l=1}^{|\mathcal{N}_i|}\exp(\phi_w^{(k)}(\mathbf{x}_i,\mathbf{z}_l))}, \\  & \phi_w^{(k)}(\mathbf{x},\mathbf{z}) =  \langle\text{FC}_{\theta_{xa}^{(k)}}(\mathbf{x}), \text{FC}_{\theta_{za}^{(k)}}(\mathbf{z})\rangle.
	\end{aligned}
	\label{eq:attention-based}
	\end{equation}
	\vskip -0.1in
	Here, $K$ is the number of attention heads. $w^{(k)}_{i, j}$ is the $k$th attentional weights between the center node $i$ and the neighboring node $j$, which is generated by applying a softmax to the dot product values. $\theta_{xa}^{(k)}$, $\theta_{za}^{(k)}$ and $\theta_{v}^{(k)}$ are the parameters of the $k$th head for computing the query, key and value vectors, which have dimensions of $d_a$, $d_a$ and $d_v$ respectively. The $K$ attention outputs are concatenated with the input vector and pass to an output fully-connected layer parameterized by $\theta_o$ to get the final output $\mathbf{y}_i$, which has dimension $d_o$. 
	The difference between our aggregator and that in GAT~\citep{velivckovic2017graph} is that we have adopted the key-value attention mechanism and the dot product attention while GAT does not compute additional value vectors and uses a fully-connected layer to compute $\phi_w^{(k)}$.

	\begin{figure}[bt!]
		\centering
		\includegraphics[width=0.45\textwidth]{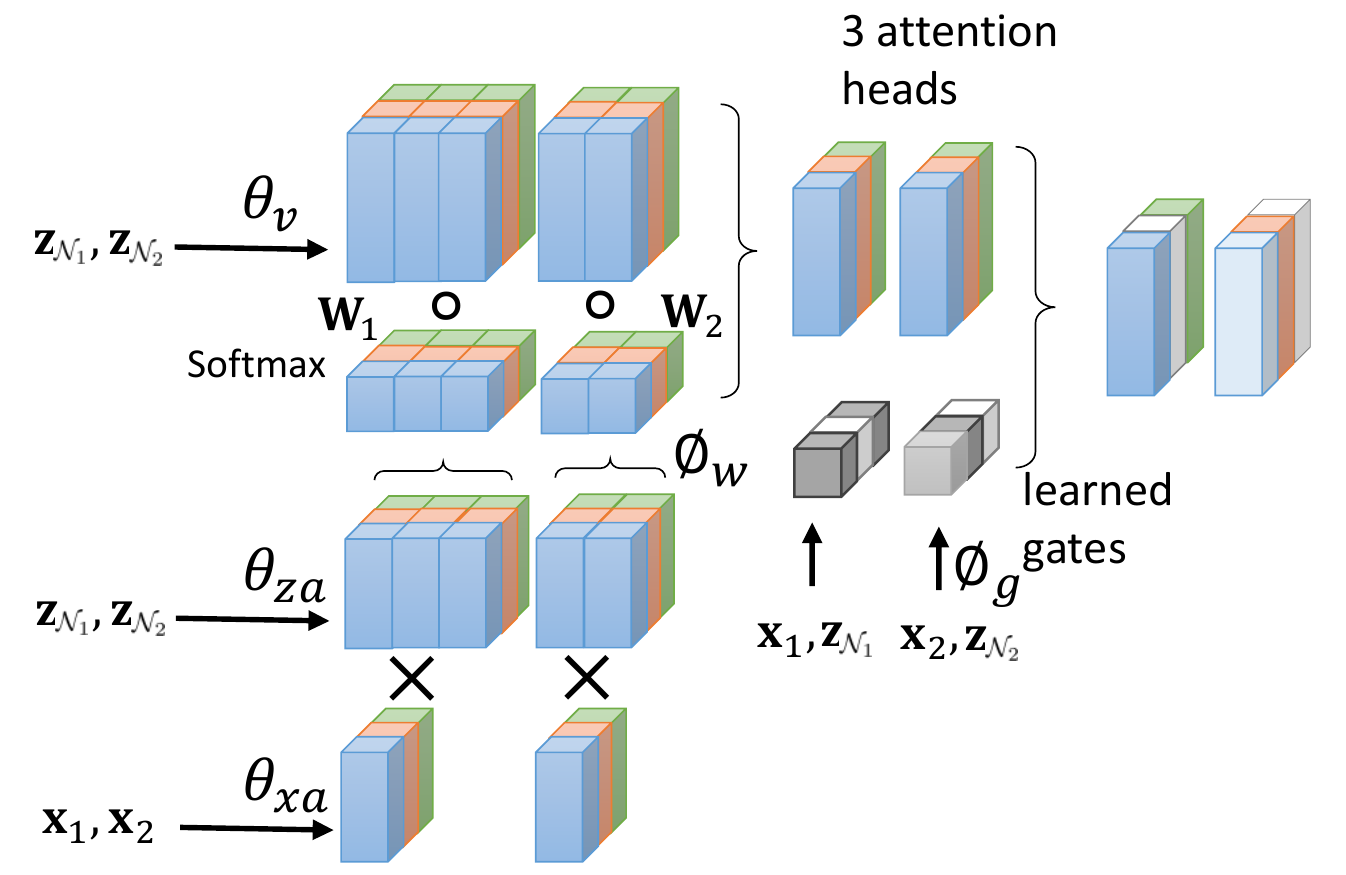}
		\caption{Illustration of a three-head gated attention aggregator with two center nodes in a mini-batch. $|\mathcal{N}_1|=3$ and $|\mathcal{N}_2|=2$ respectively. Different colors indicate different attention heads. Gates in darker color stands for larger values. (Best viewed in color)}
		\label{fig:GaAN}
	\end{figure}
	
	\begin{figure*}[tb!]
		\centering
		\begin{subfigure}[b]{0.213\textwidth}
			\centering
			\includegraphics[width=\textwidth]{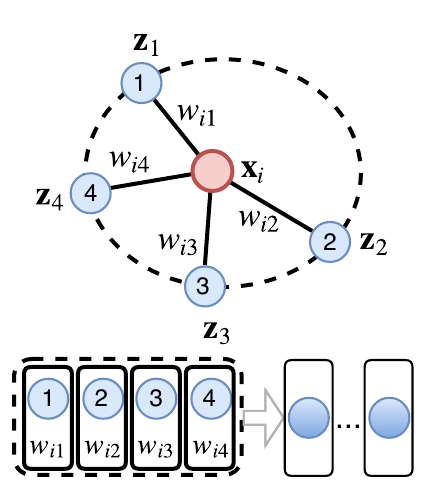}
			\caption{ Attention Aggregator}
			\label{fig:attention}
		\end{subfigure}
		~
		\begin{subfigure}[b]{0.29\textwidth}
			\centering
			\includegraphics[width=\textwidth]{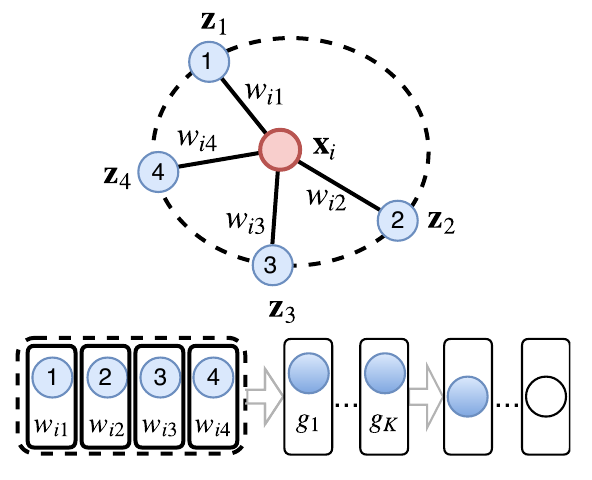}
			\caption{Gated Attention Aggregator}
			\label{fig:gated_attention}
		\end{subfigure}
		~
		\begin{subfigure}[b]{0.185\textwidth}
			\centering
			\includegraphics[width=\textwidth]{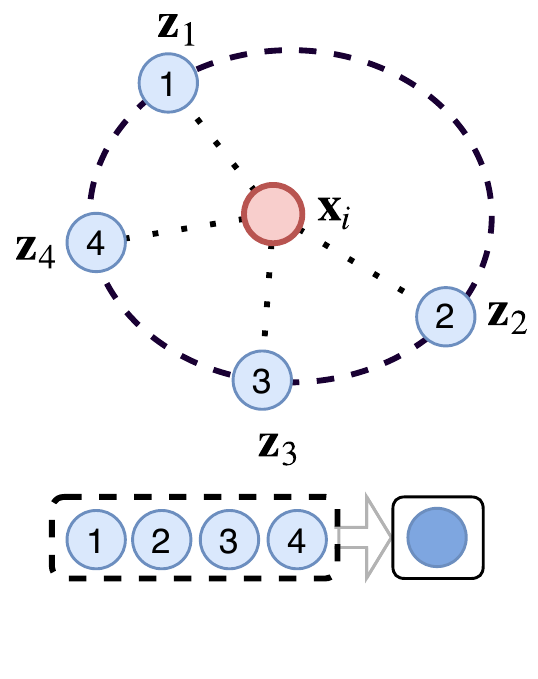}
			\caption{Pooling Aggregator}
			\label{fig:pooling}
		\end{subfigure}
		~
		\begin{subfigure}[b]{0.23\textwidth}
			\centering
			\includegraphics[width=0.86\textwidth]{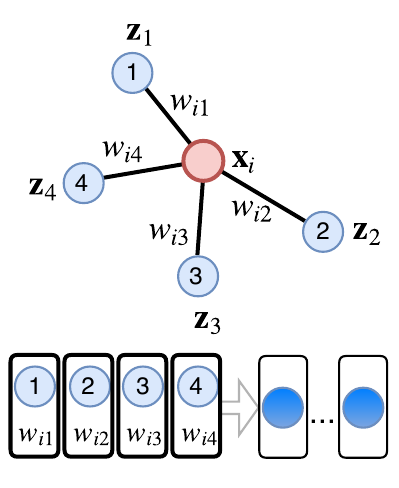}
			\caption{ Pairwise Sum Aggregator}
			\label{fig:pairwise}
		\end{subfigure}
		\caption{Comparison of different graph aggregators. The aggregators are drawn for only one aggregation step. The nodes in red are center nodes and the nodes in blue are neighboring nodes. The bold black lines between the center node and neighbor nodes indicate that a learned pairwise relationship is used for calculating the relative importance. The oval in dash line around the neighbors means the interaction among neighbors is utilized when determining the weights. (Best viewed in color)}
		\label{fig:aggregators}
	\end{figure*}
	
	\subsection{GATED ATTENTION AGGREGATOR}
	While the multi-head attention aggregator has the ability to explore multiple representation subspaces between the center node and its neighborhoods, not all of these subspaces are equally important; some subspaces may not even exist for certain nodes. Feeding the output of an attention head that captures a useless representation can mislead the model's final prediction.

	Therefore, we compute an additional soft gate between 0 (low importance) and 1 (high importance) to assign different importance to each head. In combination with the multi-head attention aggregator, we get the formulation of the gated attention aggregator:
	\begin{equation}
	\begin{aligned}
	&\mathbf{y}_i = \text{FC}_{\theta_o}(\mathbf{x}_i \oplus \bigparallel_{k=1}^K (g_i^{(k)} \sum_{j\in \mathcal{N}_i}w^{(k)}_{i, j}\text{FC}^h_{\theta_{v}^{(k)}}(\mathbf{z}_j))),\\
	&\mathbf{g}_i = [g_i^{(1)},..., g_i^{(K)}] = \psi_{g}(\mathbf{x}_i, \mathbf{z}_{\mathcal{N}_i}),
	\end{aligned}
	\end{equation}
	where $g_i^{(k)}$ is a scalar, the gate value of the $k$th head at node $i$. To make sure adding gates will not introduce too many additional parameters, we use a convolutional network $\psi_{g}$ that takes the center node and neighboring node features to generate the gate values. All the other parameters have the same meanings as in Eqn.~\eqref{eq:attention-based}.
	
	There are multiple possible designs of the $\psi_g$ network. In this paper, we combine average pooling and max pooling to construct the network. The detailed formula is given below:
	\vskip -0.1in
	\begin{equation}
	\begin{aligned}
	& \mathbf{g}_i = \text{FC}_{\theta_{g}}^\sigma( \mathbf{x}_i \oplus \max_{j\in\mathcal{N}_i}(\{\text{FC}_{\theta_{m}}(\mathbf{z}_j)\}) \oplus \frac{\sum_{j \in \mathcal{N}_i} \mathbf{z}_j}{\abs{\mathcal{N}_i}}).\\
	\end{aligned}
	\end{equation}
	\vskip -0.1in
	Here, $\theta_{m}$ maps the neighbor features to a $d_m$ dimensional vector before taking the element-wise max and $\theta_{g}$ maps the concatenated features to the final $K$ gates. By setting a small $d_m$, the subnetwork for computing the gate will have negligible computational overhead. A visual illustration of GaAN aggregator's structure can be found in Figure~\ref{fig:GaAN}. Also, we compare the general structures of the multi-head attention aggregator and the gated attention aggregator in Figure~\ref{fig:attention} and Figure~\ref{fig:gated_attention}.

	\subsection{OTHER GRAPH AGGREGATORS}
	\label{sec:existing_aggregator}
	
	Most previous graph aggregators except attention-based aggregators can be summarized into two general categories: graph pooling aggregators and graph pairwise sum aggregators. In this section, we first describe these two types of aggregators and then explain their relationship with the attention-based aggregator. Finally, we give a list of the baseline aggregators other than the multi-head attention aggregator used in the paper.

	\textbf{Graph pooling aggregators}\quad
	The main characteristic of graph pooling aggregators is that they do not consider the correlation between neighboring nodes and the center node. Instead, neighboring nodes' features are directly aggregated and the center node's feature is simply concatenated or added to the aggregated vector and then passed through an output function $\phi_o$:
	\begin{equation}
	\begin{aligned}
	\mathbf{y}_i  &= \phi_o(\mathbf{x}_i \oplus \text{pool}_{j \in \mathcal{N}_i}(\phi_v(\mathbf{z}_j))).
	\end{aligned}
	\end{equation}
	Here, the projection function $\phi_v$ and the output function $\phi_o$ can be a single fully-connected layer and the pool($\cdot$) operator can be average pooling, max pooling or sum pooling..
	
	The majority of existing graph aggregators are special cases of the graph pooling aggregators. Some models only integrate the node features of neighborhoods~\citep{duvenaud2015convolutional,kipf2017semi,hamilton2017inductive}, while others integrated edge features as well~\citep{atwood2016diffusion,fout2017protein,schutt2017quantum}. In Figure~\ref{fig:pooling}, we illustrate the architecture of the graph pooling aggregators.

	\textbf{Graph pairwise sum aggregators}\quad
	Like attention-based aggregators, graph pairwise sum aggregators also aggregate the neighborhood features by taking $K$ weighted sums. The difference is that the weight between node $i$ and its neighbor $j$ is not related to the other neighbors in $\mathcal{N}_i$. The formula of graph pairwise sum aggregator is given as follows:
	\begin{equation}
	\begin{aligned}
	\mathbf{y}_i  &= \phi_o(\mathbf{x}_i \oplus \bigparallel_{k=1}^K  \sum_{j\in \mathcal{N}_i}w^{(k)}_{i, j} \phi_v^{(k)}(\mathbf{z}_j)),\\
	w_{i, j}^{(k)} &= \phi_w^{(k)} (\mathbf{x}_i, \mathbf{z}_j).
	\end{aligned}
	\end{equation}
	Here, $w_{i,j}^{(k)}$ is only related to the pair $\mathbf{x}_i$ and $\mathbf{z}_j$ while in attention-based models $w_{i, j}^{(k)}$ is related to features of all neighbors $\mathbf{z}_{\mathcal{N}_i}$. 
	Models like the adaptive forget gate strategy in \emph{Graph LSTM}~\citep{liang2016semantic} and MoNet~\citep{monti2017geometric} employed pairwise sum aggregators with a single head or multiple heads. In Figure~\ref{fig:pairwise},  we illustrate the architecture of the graph pairwise sum aggregators.
	
	\textbf{Baseline aggregators} \quad To fairly evaluate the effectiveness of GaAN against previous work, we choose two representative aggregators in each category as baselines:
	\begin{itemize}
		\item \textbf{Avg. pooling}: $\mathbf{y}_i  = \text{FC}_{\theta_o}(\mathbf{x}_i \oplus \text{pool}^{\text{avg}}_{j \in \mathcal{N}_i}(\text{FC}^h_{\theta_{v}}(\mathbf{z}_j))).$
		\item \textbf{Max pooling}: $\mathbf{y}_i  = \text{FC}_{\theta_o}(\mathbf{x}_i \oplus \text{pool}^{\max}_{j \in \mathcal{N}_i}(\text{FC}^h_{\theta_{v}}(\mathbf{z}_j))).$
		\item \textbf{Pairwise} + \textbf{sigmoid}:
		\begin{equation}
		\nonumber
		\begin{aligned}
		\mathbf{y}_i  &= \text{FC}_{\theta_o}(\mathbf{x}_i \oplus \bigparallel_{k=1}^K \sum_{j\in \mathcal{N}_i} w_{i, j}^{(k)} \text{FC}^h_{\theta_{v}^{(k)}}(\mathbf{z}_j)),\\
		w_{i, j}^{(k)} &= \frac{1}{\abs{\mathcal{N}_i}} \sigma(\langle\text{FC}_{\theta_{xa}^{(k)}}(\mathbf{x}_i), \text{FC}_{\theta_{za}^{(k)}}(\mathbf{z}_j)\rangle).
		\end{aligned}
		\end{equation}
		\item \textbf{Pairwise} + \textbf{tanh}:  Replace the sigmoid activation in \textbf{Pairwise} + \textbf{sigmoid} to tanh.
	\end{itemize}
	
	\section{INDUCTIVE~NODE~CLASSIFICATION}
	\subsection{MODEL}
	\label{sec:inductive_model}
	In the inductive node classification setting, every node is assigned one or multiple labels. During training, the validation and testing nodes are not observable and the goal is to predict the labels of the unseen testing nodes. Our approach follows that of~\citep{hamilton2017inductive}, where a mini-batch of nodes are sampled on each iteration during training and multiple layers of graph aggregators are stacked to compute the predictions.
	
	With a stack of $M$ layers of graph aggregators, we will first sample a mini-batch of nodes $\mathcal{B}_0$ and then recursively expand $\mathcal{B}_\ell$ to be $\mathcal{B}_{\ell+1}$ by sampling the neighboring nodes of $\mathcal{B}_\ell$. After $M$ sampling steps, we can get a hierarchy of node batches: $\mathcal{B}_1,..., \mathcal{B}_M$. The node representations, which are initialized to be the node features, will be aggregated in reverse order from $\mathcal{B}_M$ to $\mathcal{B}_0$. The representations of the last layer, i.e., the final representations of the nodes in ${\mathcal{B}_0}$, are projected to get the output. We use the sigmoid activation for multi-label classification and the softmax activation for multi-class classification. Also, we use the cross-entropy loss to train the model.

	\begin{table}[tb!]
		\centering
		\caption{Effect of the merge operation. Both methods sample a maximum of 15 neighborhoods without replacement for three recursive steps on the Reddit dataset. We start from 512 seed nodes. The total number of nodes after the $l$th sampling step is denoted as $\abs{\mathcal{B}_\ell}$. The sampling process is repeated for ten times and the mean is reported.}
		\begin{tabular}{|r | rrrr|}
			\hline
			Strategy/Sample Step  & $\abs{\mathcal{B}_0}$ & $\abs{\mathcal{B}_1}$ & $\abs{\mathcal{B}_2}$ & $\abs{\mathcal{B}_3}$ \\
			\hline
			Sample without merge & 512 & 7.8K & 124.4K & 1.9M   \\ 
			Sample and merge     & 512 & 7.5K & 70.7K  & 0.2M    \\  
			\hline
		\end{tabular}
		\label{tab:sample_and_merge}
	\end{table}
	A naive sampling algorithm is to always sample all neighbors. However, it is not practical on large graphs because the memory complexity is $O(\abs{\mathcal{V}})$ and the time complexity is $O(\abs{\mathcal{E}})$, where $\abs{\mathcal{V}}$ and $\abs{\mathcal{E}}$ are the total number of nodes and edges. Instead, similar to GraphSAGE, we only sample a subset of the neighborhoods for each node. In our implementation, at the $\ell$th sampling step, we sample $\min(\abs{\mathcal{N}_i}, S_\ell)$ neighbors without replacement for the node $i$, where $S_\ell$ is a hyperparameter that controls the maximum number of sampled neighbors at the $\ell$th step. Moreover, to improve over GraphSAGE and further reduce memory cost, we merge repeated nodes that are sampled from different seeds' neighborhoods within each mini-batch. This greatly reduces the size of $\mathcal{B}_\ell$s as shown in Table~\ref{tab:sample_and_merge}.
	
	Note that $\min(\abs{\mathcal{N}_i}, S_\ell)$ is not the same for all the nodes $i$. Rather than padding the sampled neighborhood set to the same size, we implemented new GPU kernels that directly operate on inputs with variable lengths to accelerate computations.
	
	\subsection{EXPERIMENTAL SETUP}\label{sec:exp}
	
	We performed a thorough comparison of GaAN with the state-of-the-art models, five aggregator-based models in our framework and a two-layer fully connected neural network on the PPI and Reddit datasets~\citep{hamilton2017inductive}. The five baseline aggregators include the multi-head attention aggregator, two pooling based aggregators and two pairwise sum based aggregators mentioned in Section~\ref{sec:existing_aggregator}. We also conducted comprehensive ablation analysis of these two datasets.

	The PPI dataset was collected from the molecular signatures database~\citep{subramanian2005gene}. Each node represents a protein and edges represent the interaction between proteins. Labels represent the cellular functions of each protein from gene ontology. This dataset contains 24 sub-graphs, with 20 in the training set, two in the validation set, and two in the testing set. 
	Reddit is an online discussion forum where users can post and discuss contents on different topics. Each node represents a post and two nodes are connected if they are commented by the same user. 
	The labels indicate which community a post belongs to. Detailed statistics of the datasets are listed in Table~\ref{table:nc_dataset}.
	
	\begin{table}[!tb]
		\centering
		\caption{Datasets for inductive node classification. `multi' stands for multilabel classification and `single' otherwise.}
		\vskip -0.1in
		\begin{tabular}{r  rrrr }
			\hline
			\textbf{Data}  & \#\textbf{Nodes} & \#\textbf{Edges} & \#\textbf{Fea} & \#\textbf{Classes} \\
			\hline \hline
			PPI       & 56.9K  & 806.2K   & 50  & 121(multi)  \\  
			Reddit    & 233.0K & 114.6M & 602 & 41(single)   \\ \hline 
		\end{tabular}
		\label{table:nc_dataset}
	\end{table}
	
	\subsection{MODEL ARCHITECTURES AND IMPLEMENTATION DETAIL}
	The GaAN and other five aggregator-based networks are stacked with two graph aggregators. Each aggregator is followed by the LeakyReLU activation with negative slope equals to 0.1 and a dropout layer with dropout rate set to be 0.1. The output dimension $d_o$ of all layers are fixed to be 128 except when we compare the relative performance with different output dimensions. To keep the number of parameters comparable for the multi-head models with a different number of heads, we fix the product of the dimension of the value vector and the number of heads, i.e., $d_v \times K$ to be the same when evaluating the effect of varying the number of heads. Also, the hyperparameters of the first and the second layer are assumed to be the same if no special explanation is given.

	In the PPI experiments, both pooling aggregators have $d_v = 512$, where $d_v$ means the dimensionality of the value vector projected by $\theta_v$. For the pairwise sum aggregators, the dimension of the keys $d_a$ is set to be 24, $d_v=64$ and $K=8$.
	For both GaAN and the multi-head attention based aggregator, $d_a$ is set to be 24 and the product $d_v \times K$ is fixed to be 256. For GaAN, we set $d_m$ to be 64 in the gate-generation network. Also, we use the entire neighborhoods in the mini-batch training algorithm.
	
	In the Reddit experiments, both pooling aggregators have $d_v = 1024$. For the pairwise sum aggregators, $d_a=32$, $d_v=256$ and $K=8$. For the attention based aggregators, $d_a$ is set to be 32 and $d_v \times K$ is fixed to be 512. We set the gate-generation network in GaAN to have $d_m = 64$. Also, the number of heads is fixed to 1 in the first layer for both attention-based models. The maximum number of sampled neighbors in the first and second sampling steps are denoted as $S_1$ and $S_2$ and are respectively set to be 25 and 10 in the main experiment. In the ablation analysis, we also look at the performance when setting them to be (50, 20), (100, 40) and (200, 80).
	
	To illustrate the effectiveness of incorporating graph structures, we also evaluate a two-layer fully-connected neural network with the hidden dimension of 1024 and ReLU activation.
	
	\begin{table}[!tb]
		\centering
		\caption{Summary of different models' test micro F1 scores in the inductive node classification task. In the first block, we include the best-reported results in the previous papers. In the second block, we report the results obtained by our models. For the PPI dataset, we do not use any sampling strategies. For the Reddit dataset, we use the maximum number sampling strategy with $S_1$=25 and $S_2$=10.}
		\begin{tabular}{l r r}
			\hline 
			Models / Datasets & PPI &  Reddit\\
			\hline \hline
			GraphSAGE~{\tiny \citep{hamilton2017inductive}}  & (61.2)\footnote{The performance reported in the paper is relatively low because the author has not trained their model into convergence. Also, it is not fair to compare it with the other scores because it uses the sampling strategy while the others have not.} & 95.4 \\ 
			GAT~{\tiny \citep{velivckovic2017graph}}  & 97.3 $\pm$ 0.2 & - \\ 
			Fast GCN~{\tiny \citep{chen2018fastgcn}} & - &93.7  \\  \hline
			2-Layer FNN   & 54.07$\pm$0.06 &73.58$\pm$0.09 \\ 
			Avg. pooling  &96.85$\pm$0.19 &95.78$\pm$0.07 \\ 
			Max pooling   &98.39$\pm$0.05 &95.62$\pm$0.03 \\
			Pairwise+sigmoid & 98.39$\pm$0.05 &  95.86$\pm$0.08 \\
			Pairwise+tanh    & 98.32$\pm$0.18 & 95.80$\pm$0.03 \\ 
			Attention-only   & 98.46$\pm$0.09 &96.19$\pm$0.07 \\ 
			GaAN  & \textbf{98.71$\pm$0.02} & \textbf{96.36$\pm$0.03}  \\ \hline 
		\end{tabular}
		\label{table:result_node_classifation}
	\end{table}
	
	\begin{table*}[!ht]
		\centering
		\caption{Comparison of the test F1 score on the Reddit and PPI datasets with different sampling neighborhood sizes and attention head number $K$. $S_1$ and $S_2$ are the maximum number of sampled neighborhoods in the 1st and 2nd sampling steps. `all' means to sample all the neighborhoods.}
		\begin{tabular}{l |r m{1.2cm}p{0.01cm} m{1.2cm}p{0.01cm} m{1.2cm}p{0.01cm} m{1.2cm}p{0.01cm} | r c }
			\hline
			\multirow{3}{*}{Models} &  \multicolumn{9}{c}{Reddit} &\multicolumn{2}{c}{PPI} \\
			& \multirow{2}{*}{\#Param}  & \multicolumn{1}{c}{$S_1,S_2$} & & \multicolumn{1}{c}{$S_1,S_2$} & & \multicolumn{1}{c}{$S_1,S_2$} & & \multicolumn{1}{c}{$S_1,S_2$} & & \multirow{2}{*}{\#Param}  & \multicolumn{1}{c}{$S_1, S_2$} \\
			\cline{3-3}  \cline{5-5} \cline{7-7} \cline{9-9} \cline{12-12}
			&  &  \multicolumn{1}{c}{25,10}  && \multicolumn{1}{c}{50,20}   && \multicolumn{1}{c}{100,40}   && \multicolumn{1}{c}{200,80} && & all, all \\
			\hline \hline
			2-Layer FNN & 1.71M  & 73.58$\pm$0.09  && 73.58$\pm$0.09 && 73.58$\pm$0.09 && 73.58$\pm$0.09 && 1.23M & 54.07$\pm$0.06\\ 
			Avg. pooling      & 866K & 95.78$\pm$0.07  && 96.11$\pm$0.07 && 96.28$\pm$0.05 && 96.35$\pm$0.02 && 274K & 96.85$\pm$0.19 \\ 
			Max pooling      & 866K & 95.62$\pm$0.03  && 96.06$\pm$0.09 && 96.18$\pm$0.11 && 96.33$\pm$0.04 && 274K & 98.39$\pm$0.05 \\
			Pairwise+sigmoid      & 965K & 95.86$\pm$0.08  && 96.19$\pm$0.04 && 96.33$\pm$0.05 && 96.38$\pm$0.08 && 349K & 98.39$\pm$0.05  \\
			Pairwise+tanh      & 965K & 95.80$\pm$0.03  && 96.11$\pm$0.05 && 96.26$\pm$0.03 && 96.36$\pm$0.04 && 349K & 98.32$\pm$0.18  \\  \hline\hline
			Attention-only-K1  & 562K & 96.15$\pm$0.06  && 96.40$\pm$0.05 && 96.48$\pm$0.02 && 96.54$\pm$0.07 && 168K & 96.31$\pm$0.08  \\
			Attention-only-K2    & 571K & 96.19$\pm$0.07  && 96.40$\pm$0.04 && 96.52$\pm$0.02 && 96.57$\pm$0.02 && 178K & 97.36$\pm$0.08  \\
			Attention-only-K4    & 587K & 96.11$\pm$0.06  && 96.40$\pm$0.02 && 96.49$\pm$0.03 && 96.56$\pm$0.02 && 196K & 98.09$\pm$0.07  \\
			Attention-only-K8    & 620K & 96.10$\pm$0.03  && 96.38$\pm$0.01 && 96.50$\pm$0.04 && 96.53$\pm$0.02 && 233K & 98.46$\pm$0.09  \\ \hline
			GaAN-K1   & 620K & 96.29$\pm$0.05 && 96.50$\pm$0.08 && 96.67$\pm$0.04 && 96.73$\pm$0.05 && 201K & 96.95$\pm$0.09  \\
			GaAN-K2   & 629K & 96.33$\pm$0.02 && 96.59$\pm$0.02 && 96.71$\pm$0.05 && 96.82$\pm$0.05 && 211K & 97.92$\pm$0.05  \\
			GaAN-K4   & 645K & \textbf{96.36$\pm$0.03} && \textbf{96.60$\pm$0.03} && 96.73$\pm$0.04 && \textbf{96.83$\pm$0.03} && 230K & 98.42$\pm$0.02 \\
			GaAN-K8   & 678K & 96.31$\pm$0.13 && \textbf{96.60$\pm$0.02} && \textbf{96.75$\pm$0.03} && 96.79$\pm$0.08 && 267K & \textbf{98.71$\pm$0.02}  \\
			\hline
		\end{tabular}
		\label{table:result_reddit_ppi}
	\end{table*}

	We train all the aggregator-based models with Adam~\citep{kingma2014adam} and early stopping on the validation set. Besides, we use the validation set to perform learning rate decay scheduler.  
	For Reddit, before training we normalize all the features and project all the features to a hidden dimension of 256. The initial learning rate is 0.001 and gradually decreases to 0.0001 with the decay rate of $0.5$ each time the validation F1 score does not decrease in a window of 4 epochs and early stopping occurs for 10 epochs. The gradient normalization value clips no larger than 1.0. 
	For the PPI dataset, all the input features are projected to a 64-dimension hidden state before passing to the aggregators. The learning rate begins at 0.01 and decays to 0.001 with the decay rate of 0.5 if the validation F1 score does not increase for 15 epochs and stops training for 30 epochs. 
	
	The training batch size is fixed to be $512$. Also, in all experiments, we use the validation set to select the optimal hyperparameters for training. The training, validation, and testing splits are the same as that in~\citep{hamilton2017inductive}. The micro-averaged F1 score is used to evaluate the prediction accuracy for both datasets. We repeat the training five times for Reddit and three times for PPI with different random seeds and report the average test F1 score along with the standard deviation.

	\subsection{MAIN RESULTS}
	
	We compare our model with the previous state-of-the-art methods on inductive node classification. 
	This includes GraphSAGE~\citep{hamilton2017inductive}, GAT~\citep{velivckovic2017graph},
	and FastGCN~\citep{chen2018fastgcn}. 
	The GraphSAGE model used a 2-layer sample and aggregate model with a neighborhood size of $S^{(1)}=25$ and $S^{(2)}=10$ without dropout. 
	The 3-layer GAT model consisted of 4, 4 and 6 heads in the first, second and third layer respectively. Each attention head had 256 dimensions. GAT did not use neighborhood sampling, L2 regularization, or dropout. 
	The FastGCN model is a fast version of the 3-layer, 128-dimension GCN with sampled neighborhood size being 400, 100, and 400 for each layer and no sampling is done during testing.
	Table~\ref{table:result_node_classifation} summarizes all results of the state-of-the-art models as well as the models proposed in this paper. We denote the multi-head attention aggregator as `Attention-only' in the tables and figures. We find that the proposed model, GaAN, achieves the best F1 score on both benchmarks and the other baseline aggregators can also show competitive results to the state-of-the-art. We note that aggregator-based models achieve much higher F1 score than the fully-connected model, which demonstrate the effectiveness of the graph aggregators. Our max pooling and avg. pooling baselines have higher scores on Reddit than that in the original GraphSAGE paper. This mainly contributes to our usage of dropout and the LeakyReLU activation.

	\subsection{ABLATION ANALYSIS}
	We ran a quantity of ablation experiments to analyze the performance of different graph aggregators when different hyperparameters were used. We also visualized the gates of the GaAN model.
	
	\textbf{Effect of the number of attention heads and the sample size}\quad
	We compare the performance of the aggregators when a different number of attention heads and sampling strategies are used. Results are shown in Table~\ref{table:result_reddit_ppi}.
	We find that attention-based models consistently outperform pooling and pairwise sum based models with the fewer number of parameters, which demonstrates the effectiveness of the attention mechanism in this task. Moreover, GaAN consistently beats the multi-head attention model with the same number of attention heads $K$. This proves that adding additional gates to control the importance of the attention heads is beneficial to the final classification performance. From the last two row blocks of Table~\ref{table:result_reddit_ppi}, we note that increasing the number of attention heads will not always produce better results on Reddit. In contrast, on PPI, the larger the $K$, the better the prediction results. 
	
	Also, we can see steady improvement with larger sampling sizes, which is consistent with the observation in~\citep{hamilton2017inductive}.

	\begin{figure}[tb!]
		\centering
		\begin{subfigure}[b]{0.48\textwidth}
			\centering
			\includegraphics[width=0.85\textwidth]{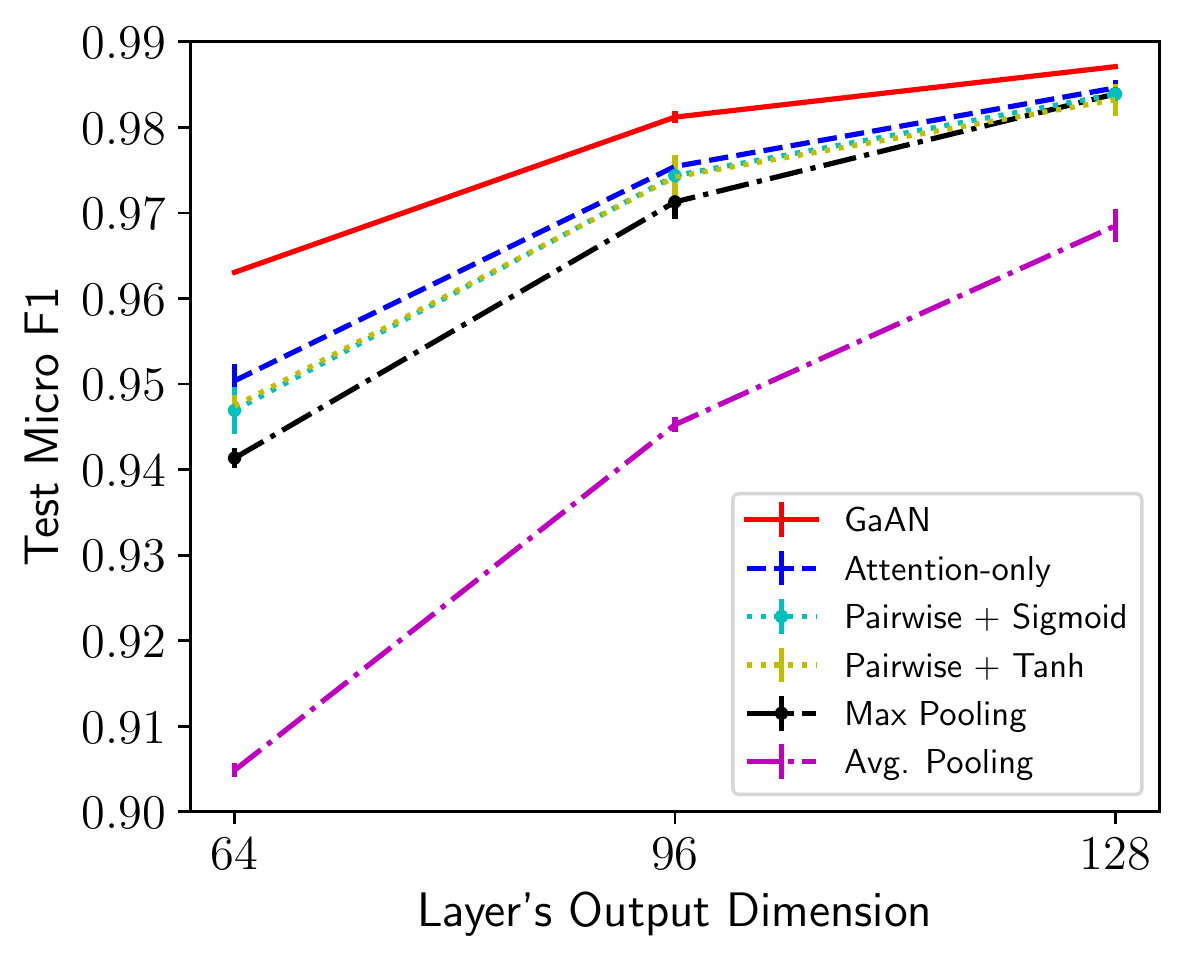}
			\caption{Performance of different models with a varying number of output dimensions on PPI.}
			\label{fig:ppi_out_dim}
		\end{subfigure}
		\begin{subfigure}[b]{0.48\textwidth}
			\centering
			\includegraphics[width=0.85\textwidth]{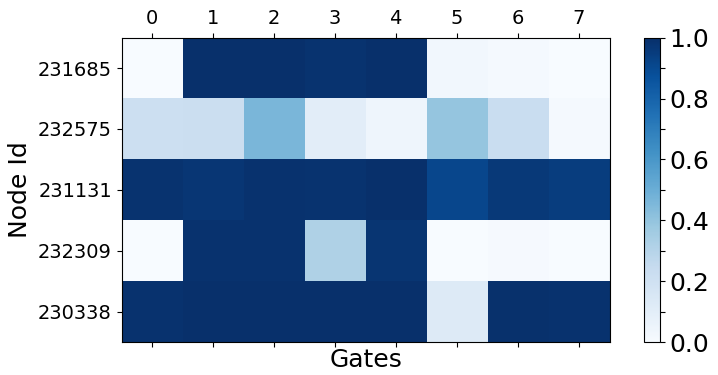}
			\caption{Visualization of 8 gate values of 5 example nodes on Reddit. Each row represents a learned gate vector for one node.}
			\label{fig:gate_vis}
		\end{subfigure}
		\caption{Ablation analysis on PPI and Reddit }
		\label{fig:ppi_ablation}
	\end{figure}
	
	\textbf{Effect of output dimensions in the PPI dataset}\quad
	We changed the output dimension to be 64, 96 and 128 in the models for training in the PPI dataset. The test F1 score is shown in Figure~\ref{fig:ppi_out_dim}. All multi-head models have $K$=8. We find that the performance becomes better for larger output dimensions and the proposed GaAN consistently outperforms the other models.
	
	\textbf{Visualization of gate values}\quad
	In Figure~\ref{fig:gate_vis}, we visualized the gate values of five different nodes output by the GaAN-K8 model trained on the Reddit dataset. It illustrates the diversity of the learned gate combinations for different nodes. In most cases, the gates vary across attention heads, which shows that the gate-generation network can be learned to assign different importance to different heads.

	\section{TRAFFIC SPEED FORECASTING}
	\subsection{GRAPH GRU}
	Following~\citep{lin2017structured}, we formulate traffic speed forecasting as a spatiotemporal sequence forecasting problem where the input and the target are sequences defined on a fixed spatiotemporal graph, e.g., the road network. To simplify notations, we denote $\mathbf{Y} = \Gamma_{\Theta}(\mathbf{X}, \mathbf{Z}; \mathcal{G})$ as applying the $\gamma$ aggregator for all nodes in $\mathcal{G}$, i.e., $\mathbf{y}_i = \gamma_{\Theta}(\mathbf{x}, \mathbf{z}_{\mathcal{N}_i})$. Based on a given graph aggregator $\Gamma$, we can construct a GRU-like RNN structure using the following equations:
	\begin{equation}
	\label{eq:graph_gru}
	\small{
		\begin{aligned}
		\mathbf{U}_t = &\sigma(\Gamma_{\Theta_{xu}}(\mathbf{X}_t, \mathbf{X}_t; \mathcal{G}) + \Gamma_{\Theta_{hu}}(\mathbf{X}_t \oplus \mathbf{H}_{t-1}, \mathbf{H}_{t-1}; \mathcal{G})),\\
		\mathbf{R}_t = &\sigma(\Gamma_{\Theta_{xr}}(\mathbf{X}_t, \mathbf{X}_t; \mathcal{G}) + \Gamma_{\Theta_{hr}}(\mathbf{X}_t \oplus \mathbf{H}_{t-1}, \mathbf{H}_{t-1}; \mathcal{G})),\\
		\mathbf{H}^\prime_t =& h(\Gamma_{\Theta_{xh}}(\mathbf{X}_t, \mathbf{X}_t; \mathcal{G})  + \mathbf{R}_t \circ \Gamma_{\Theta_{hh}}(\mathbf{X}_t \oplus \mathbf{H}_{t-1}, \mathbf{H}_{t-1}; \mathcal{G})),\\
		\mathbf{H}_t =& (1 - \mathbf{U}_t) \circ \mathbf{H}^\prime_t + \mathbf{U}_t \circ \mathbf{H}_{t-1}.
		\end{aligned}
	}
	\end{equation}
	Here, $\mathbf{X}_t \in \mathbb{R}^{\abs{\mathcal{V}} \times d_{i}}$ are the input features and $\mathbf{H}_t \in \mathbb{R}^{\abs{\mathcal{V}} \times d_o}$ are the hidden states of the nodes at the $t$th timestamp. $\abs{\mathcal{V}}$ is the total number of nodes, $d_i$ is the dimension of the input and $d_o$ is the dimension of the state. $\mathbf{U}_t$ and $\mathbf{R}_t$ are the update gate and reset gate that controls how $\mathbf{H}_t$ is calculated. $\mathcal{G}$ is the graph that defines the connection structure between different nodes.
	
	\begin{figure}[bt!]
		\centering
		\includegraphics[width=0.48\textwidth]{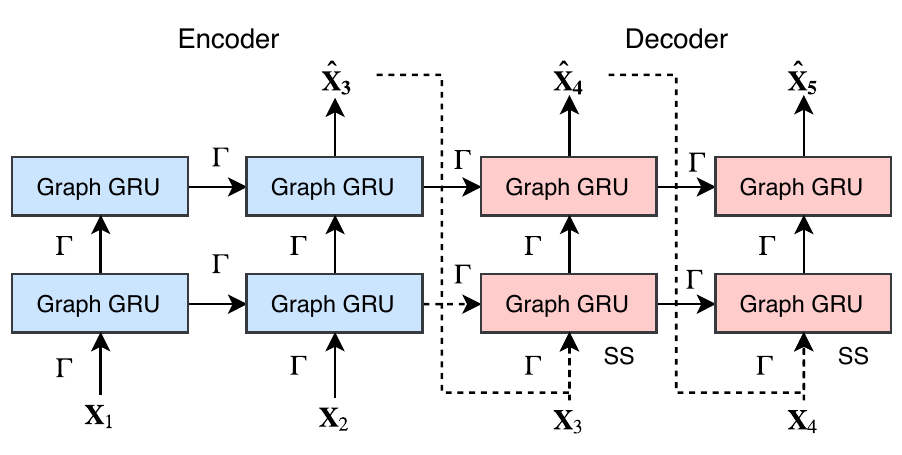}
		\caption{Illustration of the encoder-decoder structure used in the paper. We use two layers of Graph GRUs to predict a length-3 output sequence based on a length-2 input sequence. `SS' denotes the scheduled sampling step.}
		\label{fig:encoder_decoder}
	\end{figure}
	

	\begin{table*}[tb!]
		\centering
		\caption{Performance comparison of different models for traffic speed forecasting on the METR-LA dataset. Models marked with `$\dagger$' treat sensor map as a directed graph while other models convert it into an undirected graph. Scores under ``$\tau$min'' are the scores at the $\frac{\tau}{5}$th predicted frame. The last three columns contain the average scores of the 15 min, 30 min, and 60 min forecasting horizons.} 
		\vskip -0.1in
		\begin{tabular}{l m{0.55cm}m{0.55cm}m{0.65cm} p{0.01cm} m{0.55cm}m{0.55cm}m{0.65cm} p{0.01cm} m{0.55cm}m{0.55cm}m{0.65cm} p{0.01cm} m{0.55cm}m{0.55cm}m{0.65cm}}
			\hline 
			\multirow{2}{*}{Models / T}   & \multicolumn{3}{c}{15 min} && \multicolumn{3}{c}{30 min}  && \multicolumn{3}{c}{60 min} && \multicolumn{3}{c}{Average}\\
			\cline{2-4}  \cline{6-8} \cline{10-12}  \cline{14-16}  
			& {\small MAE} & {\small RMSE} & {\small MAPE} && {\small MAE} & {\small RMSE} & {\small MAPE} && {\small MAE} & {\small RMSE} & {\small MAPE} && {\small MAE} & {\small RMSE} & {\small MAPE}\\
			\hline \hline
			FC-LSTM~{\tiny \citep{li2017graph}} & 3.44 & 6.30 & 9.6\% && 3.77 & 7.23 & 10.9\%&& 4.37 & 8.69 & 13.2\% && 3.86& 7.41& 11.2\%\\
			GCRNN~{\tiny \citep{li2017graph}}  & 2.80 & 5.51 & 7.5\% && 3.24 & 6.74 & 9.0\% && 3.81 & 8.16 & 10.9\% && 3.28 & 6.80 & 9.13\% \\
			$\text{DCRNN}^\dagger$~{\tiny \citep{li2017graph}}  & 2.77 & 5.38 & 7.3\% && 3.15 & 6.45 & 8.8\% && \textbf{3.60} & \textbf{7.60} & \textbf{10.5}\% && 3.17 & 6.48& 8.87\%\\ \hline
			Avg Pool & 2.79 & 5.42 & 7.26\% && 3.20 & 6.52 & 8.84\% && 3.69 & 7.69 & 10.73\% && 3.22 & 6.54 & 8.94\% \\
			Max Pool & 2.77 & 5.36 & 7.21\% && 3.18 & 6.45 & 8.78\% && 3.69 & 7.73 & 10.80\% && 3.21 & 6.51 & 8.93\%\\
			Pairwise + Sigmoid & 2.76 & 5.36 & 7.14\% && 3.18 & 6.46 & 8.72\% && 3.70 & 7.73 & 10.77\% && 3.22 & 6.52 & 8.88\%\\
			Pairwise + Tanh & 2.76 & 5.34 & 7.14\% && 3.18 & 6.46 & 8.73\% && 3.70 & 7.73 & 10.73\% && 3.21 & 6.51 & 8.87\%\\
			Attention-only      & 2.74 & 5.33 & 7.09\% && 3.16 & 6.45 & 8.69\% && 3.67 & 7.61 & 10.77\% && 3.19 & 6.49 & 8.85\% \\
			GaAN      & \textbf{2.71} & \textbf{5.24} & \textbf{6.99}\% && \textbf{3.12} & \textbf{6.36} & \textbf{8.56}\% && 3.64 & 7.65 & 10.62\% && \textbf{3.16} & \textbf{6.41} & \textbf{8.72}\% \\
			\hline
		\end{tabular}
		\label{table:result_traffic_speed}
	\end{table*}
	
	We refer to this RNN structure as \emph{Graph GRU} (GGRU). GGRU can be used as the basic building block for RNN encoder-decoder structure~\citep{lin2017structured} to predict the future K steps of traffic speeds in the sensor network $\mathbf{\hat{X}}_{J+1}, \mathbf{\hat{X}}_{J+2}, ..., \mathbf{\hat{X}}_{J + K}$ based on the previous $J$ steps of observed traffic speeds $\mathbf{X}_1, \mathbf{X}_2, ..., \mathbf{X}_J$. In the decoder, we use the scheduled sampling
	technique described in~\citep{lin2017structured}. Figure~\ref{fig:encoder_decoder} illustrates the encoder-decoder structure in the paper.
	When attention-based aggregators are used, i.e., the multi-head attention aggregator or our GaAN aggregator, the connection structure in the recurrent step will also be learned based on the attention process. This can be viewed as an extension of \emph{Trajectory GRU} (TrajGRU)~\citep{shi2017deep} on irregular, graph-structured data.
	\begin{table}[!tb]
		\centering
		\caption{The Dataset used for traffic speed forecasting.}
		\begin{tabular}{r ccc}
			\hline
			\textbf{Data}  & \#\textbf{Nodes} & \#\textbf{Edges}  & \#\textbf{Timestamps} \\
			\hline \hline
			METR-LA   & 207    & 1,515  & 34,272 \\\hline
		\end{tabular}
		\label{table:st_dataset}
	\end{table}
	
	\subsection{EXPERIMENTAL SETUP}
	To evaluate the proposed GGRU model on traffic speed forecasting, we use the METR-LA dataset from~\citep{li2017graph}. The dataset contains traffic information of the highways of Los Angeles County. The nodes in the dataset represent sensors measuring traffic speed and edges denote proximity between sensor pairs measured by road network distance. The sensor speeds are recorded every five minutes. Complete dataset statistics are given in Table~\ref{table:st_dataset}.
	
	We follow~\citep{li2017graph}'s way to split the dataset. The first 70\% of the sequences are used for training, the middle 10\% are used for validation and the final 20\% are used for testing. We also use the same evaluation metrics as in~\citep{li2017graph} for evaluation, including \emph{Mean Absolute Error} (MAE), \emph{Root Mean Squared Error} (RMSE), and \emph{Mean Absolute Percentage Error} (MAPE). A sequence of length 12 is used as the input to predict the future traffic speed in one hour (12 steps).
	
	\subsection{MAIN RESULTS}
	
	We compare six variations of the proposed GGRU architecture with three baseline models, including fully-connected LSTM, GCRNN, and DCRNN~\citep{li2017graph}. We use the same set of six aggregators as in the inductive node classification experiment to construct the GGRU and we use two layers of GGRUs with the state dimension of 64 both in the encoder and the decoder. For attention based models, we set $K=4$, $d_a=16$ and $d_v=16$. For GaAN, we set $d_m=64$ and only use max pooling in the gate-generation network. For pooling based aggregators, we set $d_v=128$. For pairwise sum aggregators, we set $K=4$, $d_a=32$, and $d_v=16$.
	
	Since the road map is directed and our model does not deal with edge information, we first convert the road map into an undirected graph and use it as the $\mathcal{G}$ in Eqn.~\eqref{eq:graph_gru}. All models are trained by minimizing MAE loss with Adam optimizer. The initial learning rate is set to 0.001 and the batch-size is 64. We use the same scheduled sampling strategy as in~\citep{li2017graph}. 
	Table 1 shows the comparison of different approaches for 15 minutes, 30 minutes and 1 hour ahead
	forecasting on both datasets.
	
	The scores for 15 minutes, 30 minutes, and 1 hour ahead forecasting as well as the average scores over three forecasting horizons are shown in Table~\ref{table:result_traffic_speed}. 
	For the average score, we can see that the proposed GGRU models consistently give better results than GCRNN, which also models the traffic network as an undirected graph. Moreover, the GaAN based GGRU model, which does not use edge information, achieves higher accuracy than DCRNN, which uses edge information in the road network.

	\section{CONCLUSION AND FUTURE WORK}
	
	We introduced the GaAN model and applied it to two challenging tasks: inductive node classification and traffic speed forecasting. GaAN beats previous state-of-the-art algorithms in both cases. In the future, we plan to extend GaAN by integrating edge features and processing massive graphs with millions or even billions of nodes. 
	Moreover, our model is not restricted to graph learning. A particularly exciting direction for future work is to apply GaAN to natural language processing tasks like machine translation.
	
	\newpage

	
	\bibliography{graph}

\end{document}